# Generalizable Natural Language Processing Framework for Migraine Reporting from Social Media

Yuting Guo, MS[1], Swati Rajwal[2], Sahithi Lakamana, MS[1]
Chia-Chun Chiang, MD[3], Paul C. Menell, MD[3], Adnan H. Shahid, MBBS, MCh[3], Yi-Chieh Chen, Pharm D, BCPS[4], Nikita Chhabra, DO[5], Wan-Ju Chao, MS[6], Chieh-Ju Chao, MD[7], Todd J. Schwedt, MD[5]
Imon Banerjee, PhD[8,9], Abeed Sarker, PhD[1]

[1]Department of Biomedical Informatics, Emory University, Atlanta, GA, 30322;
[2]Department of Computer Science, NSUT, India, 110078
[3]Department of Neurology, Mayo Clinic, Rochester, MN
[4] Department of Pharmacy Services, Mayo Clinic Health System, Austin, MN
5 Department of Neurology, Mayo Clinic, Scottsdale, AZ
6 Department of Psychology, University of North Texas, TX
7 Department of Cardiology, Mayo Clinic, Rochester, MN
8 Department of Radiology, Mayo Clinic, Rochester, AZ
9 Department of Computer Science, Arizona State University, Tempe, AZ

**Abstract**

*Migraine is a high-prevalence and disabling neurological disorder. However, information migraine management in real-world settings could be limited to traditional health information sources. In this paper, we (i) verify that there is substantial migraine-related chatter available on social media (Twitter and Reddit), self-reported by migraine sufferers; (ii) develop a platform-independent text classification system for automatically detecting self-reported migraine-related posts, and (iii) conduct analyses of the self-reported posts to assess the utility of social media for studying this problem. We manually annotated 5750 Twitter posts and 302 Reddit posts. Our system achieved an F1 score of 0.90 on Twitter and 0.93 on Reddit. Analysis of information posted by our 'migraine cohort' revealed the presence of a plethora of relevant information about migraine therapies and patient sentiments associated with them. Our study forms the foundation for conducting an in-depth analysis of migraine-related information using social media data.*

**Introduction**

Electronic health records (EHRs) encapsulate knowledge specific to patients' health, and there is now widespread use of EHR data for studying targeted health-related factors. However, EHRs have limitations in capturing knowledge about factors influencing patients' well-being, such as social determinants, behavior, lifestyle, personal interests, mental health, and self-management of health problems. This peripheral information is often captured in patient-generated data on social media, which have thus emerged as key sources of health information that captures daily information from real-world settings directly from patients.[1] Data from social media has been used to supplement the information obtained from traditional sources including electronic health records (EHR) and knowledge obtained from clinical trials.[2–4] Recent advances in data science and artificial intelligence have led to the increasing use of natural language processing (NLP) approaches for analyzing free-text, noisy, big data from social media platforms, such as Twitter and Reddit.

Given a health topic, the automatic detection and curation of relevant health data from a social network on the topic are only feasible if (i) there are sufficient numbers of subscribers who self-report their health condition/status associated with that topic, and (ii) these self-reports can be automatically distinguished from general posts on the topic. Once these posts related to a specific topic can be identified and curated, the data can be used to study population or cohort-level topics.[5] Over recent years, several studies have attempted to build targeted cohorts from social media data and then study the information posted by the cohorts to understand other health-related factors. Some of these cohort-level studies have utilized NLP to analyze information on Twitter and showed the potential to improve patient-centered outcomes. For example, a recent study from our group developed an NLP pipeline to identify patients who self-reported breast cancer on Twitter.[6] Breast cancer-related tweets were first detected using keywords and hashtags, and a machine learning classifier was trained using manually annotated data to identify posts that represented personal experiences of breast cancer (i.e., self-reports) and separate them from the many generic posts (e.g., posts raising awareness or sharing news articles about breast cancer). Qualitative analysis of tweets extracted from users who self-reported breast cancer showed a wealth of information regarding mental health concerns, medication side effects, and medication adherence, demonstrating the tremendous potential of studying patient-centered outcomes that might not otherwise be available in EHRs. Similar approaches were applied to build cohorts of patients for whom comprehensive data may not be available from EHRs or populations excluded from traditional studies such as clinical trials. Such studies have focused on topics including but not limited to drug safety during pregnancy,[7,8] substance use,[9,10] intimate partner violence,[11–13], and chronic stress.[14,15] Since social media data is constantly generated, once the cohort-building methods are developed, they can be deployed to continuously grow the cohort, increase the volume of collected data, and conduct long-term studies. Recent studies have also shown that demographic characteristics of the cohorts (e.g., geolocation, race, gender) can be estimated fairly accurately, based on self-reported information, once large amounts of data are collected.[16] Broadly speaking, these recent studies have helped establish social media as an important source of digital health data and have opened up opportunities to study various health topics using a different lens than traditional studies.

Building on our past work in this space, we focus on the topic of migraine in this paper. Migraine is a highly prevalent and disabling neurological disorder with a one-year prevalence of about 12% in the general population.[17] A few studies have investigated migraine-related information on social media, including a study using NLP and sentiment analysis to study migraine-associated tweets classified as either "informative" or "expressive" tweets. They found that informative tweets were more likely to demonstrate extreme sentiment when they were positive tweets as opposed to negative tweets and that expressive tweets were more likely to demonstrate extreme sentiment when they were negative tweets as opposed to positive tweets.[18] The study, however, did not explicitly attempt to filter out posts that were not based on personal experiences. From an informatics perspective, it is also unclear if the methods used in the study can be ported to other social media platforms. While social media presents an important source of migraine-related information, these research gaps need to be addressed to establish social media as a long-term and sustainable resource for migraine-related research. In this study, we aim to develop a platform-independent framework to study patient-reported outcomes related to migraine treatment posted on social media by developing a robust NLP pipeline. The study particularly focuses on the development and evaluation of the essential component of the pipeline, which is about identifying patients who self-report migraine on social media (Twitter and Reddit). Specifically, we first implemented and evaluated a supervised machine learning classifier for detecting migraine self-reports using data from Twitter. We then performed a thorough error and bias analysis of the model. We evaluated the portability of this model by evaluating its performance on manually-annotated data from the social network Reddit, which has substantially different characteristics compared to Twitter. We manually analyzed some of the data posted by this cohort to assess the types of information posted by the cohort members, including but not limited to the mention of medications for migraine. As a use-case of this powerful pipeline, we demonstrate an automated analysis of sentiments regarding the commonly used migraine medications from the tweets/posts of self-reported migraine users. To the best of our knowledge, this is the first NLP platform proposed to analyze social media posts regarding migraine from a self-identifying cohort and demonstrate a clinically relevant use case.

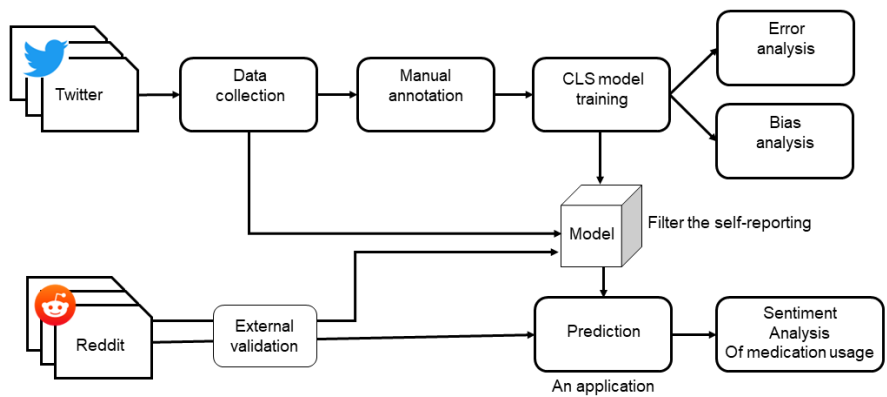

**Figure 1.** The framework of our generalizable NLP system—model development and validation on Twitter data followed by additional evaluation on Reddit posts.

## Methods
### Data collection
We collected publicly available English posts related to migraine and the user metadata from Twitter using its academic streaming application programming interface (API). We used the keyword "migraine" and the generic and brand names of migraine-specific medications, including sumatriptan (Imitrex), rizatriptan (Maxalt), naratriptan

(Amerge), eletriptan (Relpax), zolmitriptan (Zomig), frovatriptan (Frova), almotriptan (Axert), erenumab (Aimovig), fremanezumab (Ajovy), galcanezumab (Emgality), eptinezumab (Vyepti), ubrogepant (Ubrelvy), rimegepant(Nurtec), atogepant (Qulipta) to select migraine-related tweets preliminarily. Since medication names are often misspelled on social media, we used an automatic misspelling generator to obtain common misspellings for the medication names.[19] We collected the tweets between March 15, 2022, and June 28, 2022, accruing 343,559 migraine-related tweets. We also collected migraine-related posts from Reddit (r/migraine, r/NDPH, r/headache, r/headaches subreddit) using PRAW Reddit API, collecting a total of 3625 posts. The preprocessing of all posts includes lowercasing, normalization of numbers, usernames, URLs, hashtags, and text smileys, and adding extra marks for capital words, hashtags, and repeated letters.

*Annotation process*

As the first step, 500 tweets were labeled by five clinician annotators to determine whether the tweets were self-reports from Twitter users. The initial mean agreement between any two annotators was 0.60 (Cohen's kappa). To optimize the agreement between annotators, we developed an annotation guideline to determine whether tweets should be labeled as self-reporting migraine. The guideline details are included in the supplementary file on Google Drive[1], and the key rules are included in Table 1. After the guideline was implemented, we extracted an additional 250 tweets which the same five clinicians annotated. The mean agreement improved to 0.78 between 4 annotators but was 0.66 when adding the annotation of the 5th annotator. The lead clinician investigator and headache specialist (CC) reviewed the annotation guideline and resolved the discrepancies to ensure agreement. Finally, an additional 5000 tweets were labeled by the five annotators (1000 each person). A total of 5750 tweets labeled by clinicians were used to train the NLP model, which was then applied to Reddit posts. For the annotation of Reddit posts, the same annotation guideline was applied. A total of 303 Reddit posts were annotated separately by the five clinician annotators.

**Table 1** Summary of Annotation Guideline for identifying self-reported migraine patients. Annotators were asked to label tweets or Reddit posts as Y or N.

| Rules | Examples tweets |
|---|---|
| Using first person references such as "I" or "my" when referring to migraine-related experiences or intake of migraine-specific medications | *"couldnt get **all my work done** yesterday because of a migraine so working extra hard today, pray for me"* <br> *"rizatriptan **my beloved**"* <br> The above examples should be labeled Y |
| Evidence of past or future, but not current migraine, as long as it is referring to the own experience of a user, should be labeled "Yes" | *"The biggest health improvement since stopping hormonal birth control a year ago has been **no more migraines. I used to get them** monthly and they were particularly bad when I was on the arm implant. Now I can't even remember the last time I had one"* <br> The above example should be labeled Y |
| In some instances, the users might use "them" "they" "that" to refer to migraine. If the sentence includes first person references, it would be labeled "Yes" | *"Do you feel it when they give you migraines? I usually wake up with **that** on weekends."* <br> The above example should be labeled Y |
| Emojis should be interpreted in this study. If a tweet includes a sentence or emoji referring to the user's own feelings or experiences, those tweets should be labeled "Yes" <br> Note that links included in tweets should not be interpreted. | *high on a rizatriptan and deep throated leftover sushi. feeling:* 😵 <br><br> The above example should be labeled Y |

---

[1] https://docs.google.com/document/d/1fcVSmlgl14M5Ck67e7auwh60z0C8nDdj/edit?usp=sharing&ouid=112998194690665600455&rtpof=true&sd=true

| When tweets don't include first-person references, although it could potentially be inferring to the user's own migraine or experiences, however, as it could also potentially refer to someone else's (like their partner or close family member/friend) migraine, should be labeled No | "Who else gets migraines? Do you ever get a weird "drunk-like" feeling in your eyes when its going away/lingering?"<br>"You shouldn't be allowed to wake up with a migraine - "hey we know you have to wake up to endure the collective trauma of the world but here's another cherry on top just to make sure you're miserable!"<br>The above example should be labeled N |
|---|---|
| Advertisements, factual statements, or irrelevant statements should be labeled as No | "How is headache treated in emergency room in past decade? Opioids use reduced from more than 50% to just below 30%. Maybe use of migraine specific medications (triptans/ergots) has increased? Nope! Still less that 5% receive this therapy, our study shows https://t.co/cVVNZfW9R3"<br>"AbbVie announced that its oral CGRP receptor antagonist, atogepant, met its primary end point in the phase 3 PROGRESS trial, demonstrating a statistically significant reduction in mean MMDs compared with placebo in patients with chronic #migraine." |

### *Text classification model for identifying self-reporting of migraine*

The primary objective of this module is to develop a generalizable free-text classification model that can distinguish between posts related to migraine self-reporting from the generic posts since the posts are primarily retrieved only based on simple keywords. We experimented with transformer-based models since they have achieved state-of-the-art performance on a wide range of NLP tasks.[20,21] We investigated six transformer-based models to construct the self-report migraine classifier——RoBERTa,[21] SciBERT,[22] BioBERT,[23] BioClinicalBERT,[24] Clinical_KB_BERT,[25] and BERTweet.[26] These transformer-based models were pre-trained on large-scale text data from different domains. RoBERTa was pre-trained on the open domain, including English Wikipedia, news, books, Reddit comments, and story-like text from scratch; SciBERT was pretrained on scientific publications from scratch; BioBERT was initialized from BERT and pretrained on biomedical research papers; BioClinicalBERT was initialized from BioBERT and pretrained on clinical notes; Clinical_KB_BERT was initialized from BioBERT and pretrained on clinical notes with the knowledge base of the Unified Medical Language System (UMLS); BERTweet was pretrained on English tweets from scratch. We split the annotated Twitter data into the training, validation, and test sets at a 64-16-20 ratio. The model was trained on the training set for ten epochs and evaluated on the validation set for each epoch. The model checkpoint that achieved the best result on the validation set was selected for evaluation on the test set.

We applied the Twitter model to classify the Reddit posts without fine-tuning the Reddit data. First, we removed duplicate posts from the Reddit dataset to get a unique cohort. We evaluated precision, recall, and F1 score performance for the positive class while comparing it against the expert annotations. Reddit posts are often larger than Twitter, so we split the posts at the sentence level and run the classification model for each sentence. Finally, we integrated the post-classification if at least one sentence within the post is classified as self-reporting. We computed the 95% confidence interval of the $F_1$-score via bootstrapping.

### **Post-classification analysis**
### *Error analysis*

To explore the potential reason for misclassifications of self-reporting, we conducted a thorough error analysis on the model that achieved the best performance. We manually analyzed the contents of false positives from both Twitter and Reddit and categorized the error types.

### *Bias analysis*

In recent years, it has been studied that deep learning methods might introduce bias to the model[27]. To explore whether there exists a bias in our model, we conducted gender bias analysis and race bias analysis by testing whether changing the gender-related word or race-related word would change the model prediction. Specifically, for gender bias analysis, we changed gender-related words into words related to a different gender and vice versa. We checked if the change in the gender-related word would alter the classification result or significantly affect the word importance of the gender-related word in the tweet. Similarly, for race bias analysis, we replaced race-related words with words related to a different race and vice versa.

*Longitudinal data collection and analysis*
To study longitudinal information posted by Twitter users who self-reported suffering from migraine, we first used the best-performing classifier to automatically classify all migraine-related posts we initially collected. We then used the API to collect all the past posts of those users whose migraine-related posts were classified as positive. This provided us with many posts from migraine sufferers that we could analyze to identify the presence of other migraine-related information.

*Sentiment analysis*
On the timeline tweets from users who self-report migraine identified from the above step, we extracted tweets that included any of the medication names defined in Table 2. We performed sentiment analysis using a publicly available sentiment analysis tool named VADER[28], which can assign a sentiment score ranging from -1 (most extreme negative) to 1 (most extreme positive) for a tweet. We analyzed the sentiment distributions of the medications. To reduce the data noise, headache specialists (CC and TS) grouped the medications, and the details are shown in Table 1. We reported the frequency of occurrence of each medication group. For Twitter posts, if the same medication group is mentioned in multiple tweets/posts from the same user, the tweet/post that obtained the median sentiment score was chosen to represent the sentiment score of the user for that medication group. We also reported the mean, median, and standard deviation of the sentiment scores for each medication group. Additionally, we used the kernel density estimate (KDE) plot to visualize the sentiment distribution. In the case of external validation using Reddit posts, we used the same tool and reported results in a similar approach. We performed sentiment analysis on the Reddit posts which were predicted as self-reporting (positive) and contained the medication keywords.

**Results**
*Self-Reporting of Migraine Classification Results*
Table 3 listed the statistics of training, validation, and test set for Twitter and external test for Reddit data. Table 3 shows the quantitative performance of transformer-based classification models. We observed that RoBERTa achieved the best recall score, BioBERT and BERTweet achieved the best precision score, and BERTweet achieved the best F1 score. In general, BERTweet outperformed other models on this task. It can be attributed that the pretraining data of BERTweet is English tweets, which may help the model on the task in social media language more efficiently. We applied the optimal models (RoBERTa and BERTweet) to the Reddit dataset and reported the performance in Table 4.

**Table 2.** The expert-selected medications and their groupings.

| Medication group | Medications |
| --- | --- |
| Topiramate | Topiramate (Topamax) |
| Beta Blockers | Propranolol (Inderal), Atenolol (enormin), Metoprolol (Toprol) |
| Tricyclic antidepressants | Amitriptyline (Elavil). Nortriptyline (Pamelor) |
| OnabotulinumtoxinA | OnabotulinumtoxinA (Botox) |
| CGRP monoclonal antibodies | Erenumab (Aimovig), Galcanezumab (Emgality), Fremanezumab (Ajovy), Eptinezumab (Vyepti) |
| Gepants | Atogepant (Qulipta), Ubrogepant (Ubrelvy), Rimegepant (Nurtec) |
| Triptans | Sumatriptan (Imitrex), Rizatriptan (Maxalt), Eletriptan (Relpax), Naratriptan (Amerge), Frovatriptan (Frova), Zolmitriptan (Zomig), Almotriptan (Axert) |

| Lasmiditan | Lasmiditan (Reyvow) |
| --- | --- |
| Dihydroergotamine | Dihydroergotamine (DHE), Migranal |

**Table 3.** The statistics of training, validation, and test set – Twitter and Reddit posts.

| Dataset | Positive cases | Negative cases | Total |
| --- | --- | --- | --- |
| Training | 1034 | 2612 | 3646 |
| Validation | 272 | 647 | 919 |
| Test | 761 | 328 | 1089 |
| External Test: Reddit | 226 | 76 | 302 |

*Self-Reporting of Migraine Error analysis*
Due to a large number of Twitter posts, precision is more important than recall for this work. Given that BERTweet achieved the best precision and $F_1$-score compared with other models, we performed error analysis on the false positives predicted by BERTweet. We observed that the one type of false positive occurred when the tweet was imprecise, e.g. mentioned the word "migraine" without a clear indication if it was a self-reporting. For instance, "adulthood is preparing for migraines by taking ibuprofen the night before lmao". Another type of false positive occurred when the tweet was incomplete. For instance, *"These migraines Ain no ho"*. According to the annotation guideline, the former should be labeled as N, and the latter should be labeled Y due to the absence and presence of a first-person reference. However, it could be difficult to ascertain, even for a human annotator, whether the user had a migraine or not, which can be more difficult for the NLP model.

*Self-Reporting of Migraine Bias analysis*
We performed the manual analysis on 5% of all the tweets in our test set. Table 4 presents examples of this analysis. We observed that the change in the gender-related word or the race-related word slightly affected the word importance of the word, but the difference was trivial and did not alter the classification results. It suggests that our model is not biased towards a specific gender or race group.

**Table 4.** The classification results of 6 transformer-based models.

| Model | Precision | Recall | f1-score (95% CI) |
| --- | --- | --- | --- |
| Twitter Data | | | |
| RoBERTa | 0.84 | **0.95** | 0.89 (0.87-0.91) |
| SciBERT | 0.87 | 0.89 | 0.88 (0.85-0.90) |
| BioBERT | **0.88** | 0.89 | 0.88 (0.86-0.91) |
| BioClinicalBERT | 0.85 | 0.91 | 0.88 (0.86-0.91) |
| BERTweet | **0.88** | 0.91 | **0.90 (**0.87-0.92) |
| Clinical_KB_BERT | 0.86 | 0.91 | 0.88 (0.85-0.90) |
| External: Reddit data | | | |
| RoBERTa | 0.91 | 0.95 | 0.93 (0.91-0.95) |
| BERTweet | 0.89 | 0.90 | 0.90 (0.87-0.93) |

**Table 5.** The examples for bias analysis on the predictions of self-reported and non-self-reported migraine posts. The pharmaceutical company name is presented as "*<company>*" in the tweet and corresponding figure.

| Social Media | Example | Human annotation | Model prediction | Word importance visualization |
|---|---|---|---|---|
| Twitter | I know nearly every **girl** says this, but I always feel sick and have a tummy ache and migraine and I bet ya a tenner more than anyone else | Self-report | Self-report | Figure 2(a) |
| Twitter | <company> Real-World Study Highlights Increased Healthcare Utilization Among **Americans** with Episodic Migraine having Higher Levels of Migraine-Related Disability | Non-self-report | Non-self-report | Figure 2(b) |
| Reddit | This is Asher. **He** likes to pet my face with **his** soft toe beans when I'm in bed with a migraine. I've had about 19 headache/migraine days in the last month. I just started Aimovig on Tuesday, so I'm hoping to reduce the number of migraine days, just not the number of cuddling days. My sweet **boy**. | Self-report | Self-report | Figure 2(c) |
| Reddit | My **husband** is a teacher. Yesterday, **he** turned on the lights in **his** classroom, to which a young **woman** in **his** class visibly flinched. **He** turned off the lights again right away and **she** breathed a sigh of relief. Upon asking if **she's** okay, **she** said **she** was absent the last couple of days due to migraines. After having seen me suffer, my SO now recognises some migraine signs. I'd like to think **his** simple action of turning off the lights made this **girl's** day a little more bearable. | Non-self-report | Non-self-report | Figure 2(d) |

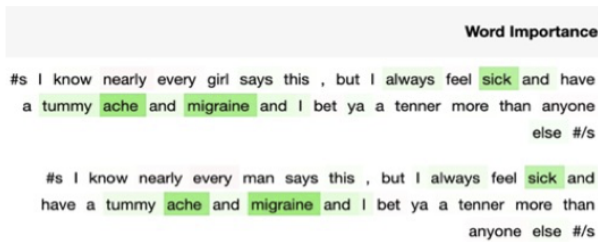

(a)

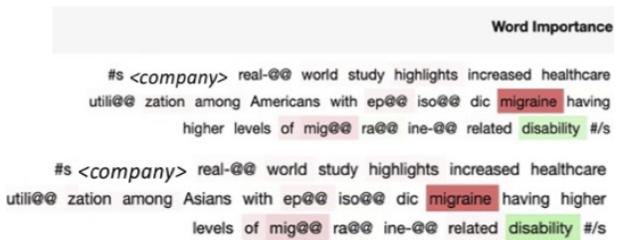

(b)

## Word Importance

#s This is Ash@@ er . He likes to pet my face with his soft toe beans when I@@ #unk m in bed with a migraine . I@@ #unk ve had about 19 head@@ ache@@ /@@ migraine days in the last month . I just started Ai@@ movi@@ g on Tuesday , so I@@ #unk m hoping to reduce the number of migraine days , just not the number of cuddling days . My sweet boy . #unk #/s

#s This is Ash@@ er . She likes to pet my face with her soft toe beans when I@@ #unk m in bed with a migraine . I@@ #unk ve had about 19 head@@ ache@@ /@@ migraine days in the last month . I just started Ai@@ movi@@ g on Tuesday , so I@@ #unk m hoping to reduce the number of migraine days , just not the number of cuddling days . My sweet girl . #unk #/s

(c)

## Word Importance

#s My husband is a teacher . Yesterday , he turned on the lights in his classroom , to which a young woman in his class visi@@ bly flin@@ ched . He turned off the lights again right away and she breath@@ ed a sigh of relief . Upon asking if she@@ 's okay , she said she was absent the last couple of days due to migraines . After having seen me suffer , my SO now recogn@@ ises some migraine signs . I@@ 'd like to think his simple action of turning off the lights made this girl@@ 's day a little more bearable . #/s

#s My wife is a teacher . Yesterday , she turned on the lights in her classroom , to which a young man in her class visi@@ bly flin@@ ched . She turned off the lights again right away and he breath@@ ed a sigh of relief . Upon asking if he@@ 's okay , he said he was absent the last couple of days due to migraines . After having seen me suffer , my SO now recogn@@ ises some migraine signs . I@@ 'd like to think her simple action of turning off the lights made this boy@@ 's day a little more bearable . #/s

(d)

**Figure 2.** The gender bias and race bias analysis examples. The green color signifies positive attention to the words while red shows negative.

*Sentiment analysis*

Table 5 shows the results of sentiment analysis for both Twitter and Reddit, and Figure 3 illustrates the sentiment distributions of the medications of which the frequency is greater than 40. In Twitter, the sentiment scores of onabotulinumtoxinA, triptans, topiramate, beta-blockers, and tricyclic antidepressants are concentrated in the 0 mean. It suggests that the sentiment for these medications tends to be neutral. On the other hand, the sentiment distributions of CGRP monoclonal antibodies and Gepants tend to be more positive. It is worth noting that the sentiment score was evaluated on the whole tweet rather than the medication mention in the tweet. It is possible that the positive sentiment in the tweet is irrelevant to the medication. For example, "The woman behind me @<user_name> didn't want her free chocolate chip cookie so she gave it to me! My day is looking up! On to my Botox injections! #EDS #lucky #winner #happy" and "Botox was approved for migraines!! Slowly but surely i'm making my symptoms manageable/not constant". In the above examples, both tweets had positive sentiments, though only the sentiment of the second tweet refers directly to the medication itself. On the other hand, the sentiment analysis on Reddit posts shows clearer positive and negative trends. For example, beta-blockers have a positive sentiment score while topiramate posts have a mean negative sentiment. However, note that conducting sentiment analysis at the medication level is a very challenging task, particularly when the data pull is smaller and requires further investigation with a larger data pull with the equal presence of posts with each medication group for making any reasonable clinical conclusion.

**Table 6.** The statistics of the medication groups.

| Medication Group | Frequency | Mean | Median | Standard deviation |
|---|---|---|---|---|
| Twitter | | | | |
| OnabotulinumtoxinA | 521 | 0.05 | 0.5 | 0.14 |
| Triptans | 174 | 0 | 0.57 | -0.05 |
| CGRP monoclonal antibodies | 103 | 0 | 0.54 | -0.01 |
| Gepants | 87 | 0.16 | 0.5 | 0.14 |
| Topiramate | 55 | 0 | 0.5 | -0.03 |
| Beta Blockers | 41 | 0 | 0.48 | 0.03 |
| **Reddit** | | | | |
| Topiramate (Topamax) | 32 | -0.17 | -0.36 | 0.71 |
| Beta Blockers | 18 | 0.39 | 0.63 | 0.64 |
| Tricyclic Antidepressants | 30 | -0.3 | -0.66 | 0.74 |
| OnabotulinumtoxinA (botox) | 41 | 0.01 | 0.18 | 0.75 |
| CGRP monoclonal antibodies | 41 | -0.07 | -0.3 | 0.77 |
| Gepants | 39 | -0.04 | 0.25 | 0.78 |
| Triptans | 64 | -0.03 | 0.06 | 0.8 |

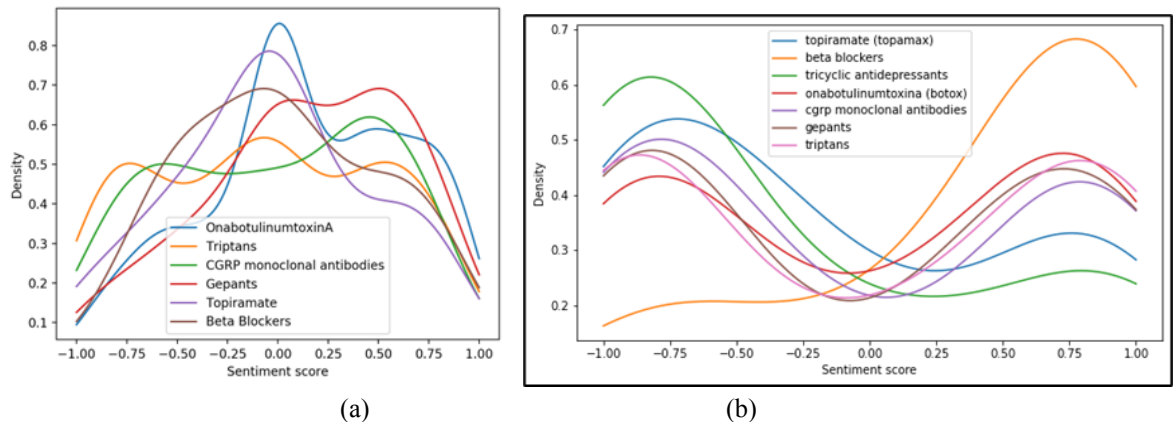

(a)          (b)

**Figure 3.** The normalized sentiment distributions of the medications: (a) Twitter and (b) Reddit.

**Conclusion**

We proposed a robust NLP framework to perform a large-scale social media study of migraine. We first trained and evaluated a machine learning classifier to automatically detect self-reports of migraine from Twitter. Being trained on only the Twitter posts, the classification framework showed generalizability towards both shorter tweets as well as for the longer Reddit posts. Moreover, our pipeline displayed minimal bias towards gender- and race-related words and may present a fair performance across the validation data. Finally, we presented a brief sentiment analysis of migraine medications as a possible use case for our system. We anticipate that our methods can be applied to collect information regarding migraine management in real-world settings and the effectiveness and tolerability of these different treatment strategies and could be applied to other topics too.